\documentclass[10pt,twocolumn,letterpaper]{article}

\usepackage{cvpr} % uncomment this for the final submission
\usepackage{times}
\usepackage{epsfig}
\usepackage{graphicx}
\usepackage{amsmath}
\usepackage{amssymb}

\usepackage{amsfonts}
\usepackage{algorithmic}
\usepackage{textcomp}
\usepackage{soul}
\usepackage{hyperref}
\usepackage{makecell}
\usepackage{multirow}
\usepackage{array}
\usepackage{caption}
\usepackage{placeins}
\usepackage{tabularx}
\usepackage{booktabs}
\usepackage{tablefootnote}
\usepackage{color}
\usepackage[table]{xcolor}
\usepackage{array}
\usepackage{cite}
\usepackage{pifont}

\usepackage{bookmark}
\bookmarksetup{
  numbered, 
  open,
}

\usepackage[capitalize]{cleveref}
\crefname{section}{Sec.}{Secs.}
\Crefname{section}{Section}{Sections}
\Crefname{table}{Table}{Tables}
\crefname{table}{Tab.}{Tabs.}


\begin{document}

%%%%%%%%% TITLE
\title{Securing Face and Fingerprint Templates in Humanitarian Biometric
Systems}

\author{
Giuseppe Stragapede$^{\ast}$\textsuperscript{1},
Sam Merrick$^{\ast}$,
Vedrana Krivoku\'ca Hahn$^{\dagger}$,
Justin Sukaitis$^{\ddagger}$,
Vincent Graf Narbel$^{\ddagger}$ \\
$^{\ast}$Simprints, $^{\dagger}$Idiap Research Institute, $^{\ddagger}$International Committee of the Red Cross (ICRC)}

% % omit the following lines up until the closing `}''.
% % Additional authors and addresses can be added with `\and'',
% \and

\maketitle
\thispagestyle{empty}

%%%%%%%%% ABSTRACT
\begin{abstract}
In humanitarian and emergency scenarios, the use of biometrics can dramatically improve the efficiency of operations, but it poses risks for the data subjects, which are exacerbated in contexts of vulnerability. To address this, we present a mobile biometric system implementing a biometric template protection (BTP) scheme suitable for these scenarios. 
After rigorously formulating the functional, operational, and security and privacy requirements of these contexts, we perform a broad comparative analysis of the BTP landscape. \textit{PolyProtect}, a method designed to operate on neural network face embeddings, is identified as the most suitable method due to its effectiveness, modularity, and lightweight computational burden. 
We evaluate \textit{PolyProtect} in terms of verification and identification accuracy, irreversibility, and unlinkability, when this BTP method is applied to face embeddings extracted using \textit{EdgeFace}, a novel state-of-the-art efficient feature extractor, on a real-world face dataset from a humanitarian field project in Ethiopia. Moreover, as PolyProtect promises to be modality-independent, we extend its evaluation to fingerprints. To the best of our knowledge, this is the first time that PolyProtect has been evaluated for the identification scenario and for fingerprint biometrics. Our experimental results are promising, and we plan to release our code\footnotemark[2].
\footnotetext[1]{\href{giuseppe@simprints.com}{giuseppe@simprints.com}}
\footnotetext[2]{\href{https://github.com/Simprints/Biometrics-SimPolyProtect}{https://github.com/Simprints/Biometrics-SimPolyProtect}}

\end{abstract}

\section{Introduction}
\label{sec:1_Introduction}
In humanitarian and emergency scenarios, biometric recognition can represent an efficient solution to verify or establish the identity of beneficiaries during the distribution of physical goods, such as medicines, food or blankets. In these contexts, IDs and access tokens are difficult to distribute, and passwords can be forgotten or shared with unentitled subjects.
On the other hand, biometrics might introduce risks for data subjects, such as impersonation \cite{otroshi2024face}, inference of personal and sensitive information \cite{terhorst2021soft}, linkage of individuals across application databases (cross-matching) \cite{unlink}, etc. These risks can be exacerbated if the contexts in which beneficiaries live makes them especially vulnerable. For example, when the Taliban took control of Afghanistan, they gained access to biometric devices left behind by the US Army, which contained data about Afghan civilians The Taliban used them to determine who had a relationship with the US Army \cite{janus, hrw}. For these reasons, the security standards upheld by humanitarian organizations are stricter, in line with the `do-no-harm' principle \cite{dphb}.  

Fortunately, the biometrics research community has invested significant efforts towards enhancing the security of biometric operations, with the establishment of the biometric template protection (BTP) research field \cite{nandakumar2015biometric, isobtp}. BTP aims to develop methods that can be applied to biometric data to produce a protected version that can be safely stored, revealing little or no information about the data subjects. In particular, a BTP method should possess the following properties \cite{isoperf}:
\textit{(i)} recognition accuracy: the incorporation of the protection method into a biometric system should not degrade the system's recognition accuracy;
\textit{(ii)} irreversibility: it should be computationally infeasible to recover the original biometric data from its protected version; 
\textit{(iii)} unlinkability and renewability: it should be possible to generate multiple distinct protected templates from the same subject's biometric data, so that the protected templates cannot be linked to each other. This would allow for the revocation and renewal of compromised templates, as well as the use of the same biometric characteristic across multiple databases, without the risk of cross-matching the data.

In this work, we propose a BTP-enhanced biometric system suitable for humanitarian use-cases. Firstly, we provide a rigorous formulation of the specific functional, operational, security and privacy requirements for the adopted BTP solution (Sec. \ref{sec:2_requirements}). The characteristics of our use-case do not impair the generality of the gathered requirements, which could be applied to other security-critical mobile applications. Secondly, we perform a thorough analysis of the entire BTP landscape (Sec. \ref{sec:3_Landscape}), which results in identifying \textit{PolyProtect}, a feature-transformation approach designed for mobile face verification \cite{polyprotect}, as the most suitable method, over well-known approaches such as homomorphic encryption (HE) or hashing-based solutions. 
Thirdly, we achieve solid experimental results (Sect. \ref{sec:6_Results}), assessing PolyProtect from the perspective of verification (obtaining improved recognition rates in comparison with the unprotected system) and identification performance, irreversibility and unlinkability, in combination with \textit{EdgeFace}, a novel state-of-the-art efficient feature extractor, on a face dataset from a field project in Ethiopia. Moreover, as PolyProtect promises to be modality-independent, we extend its evaluation to fingerprint biometrics, showing the true cross-modality potential of this BTP method. To the best of our knowledge, this is the first time that PolyProtect has been evaluated in the context of identification and for fingerprint biometrics, as well as on real-world datasets collected in humanitarian field projects.

\section{BTP System Requirements}
\label{sec:2_requirements}
\textbf{Functional Requirements}

\textit{F.1 -- Recognition Accuracy} 
There should be no degradation of the recognition performance of the protected biometric system with respect to its unprotected counterpart. 

\textit{F.2 -- Modality-Independence} 
The designed BTP solution should be applicable to all modalities. 

\textit{F.3 -- Feature Extractor-Independence} 
It should be possible to combine the BTP solution with different biometric feature extractors. 
 
\textit{F.4 -- On-device Recognition} 
The enrolment, verification and identification of subjects should take place on the device without any internet connectivity. 

\textit{F.5 -- Easy New Enrolment}
The enrolment of new subjects should not create any conflict with the running BTP solution nor with previously enrolled subjects.

\textit{F.6 -- Template Revocability and Renewability}
If the protected templates are compromised, \eg, in the case of a reported missing device, it should be possible to revoke them and reissue new protected instances.

\textit{F.7 -- Open-Source}
The BTP solution should not use closed-source or commercial solutions, relying instead on technologies released under open-source-compatible license terms.

\textbf{Operational Requirements}

\textit{O.1 -- Computational Efficiency}
Lightweight BTP solutions should be preferred due to the mobile environment (\eg, low-cost smartphones) resource constraints. 

\textit{O.2 -- Time Efficiency}
Fast BTP solutions should be preferred, \eg, time should not represent an issue for identification against a large enrolment database stored locally.

\textit{O.2 -- Offline Processing}
The BTP solution should not rely on any remotely available resource.

\textbf{Security and Privacy Requirements}

\textit{S.1 -- Irreversibility}
The adherence of the BTP solution to the irreversibility criterion should be demonstrated theoretically or empirically.

\textit{S.2 -- Unlinkability}
The adherence of the BTP solution to the unlinkability criterion should be demonstrated theoretically or empirically.

\textbf{Threat Model} 

To evaluate the security and privacy properties of a BTP method (irreversibility and unlinkability), it is first necessary to define a threat model, which characterises the type of attacker on which this analysis is based. Several threat models are defined in the ISO/IEC 30316:2018 standard on performance testing of biometric template protection schemes \cite{isoperf}. 
In our use-case, a realistic scenario is represented by an attacker stealing one or more storage devices, which can be rooted to: obtain full knowledge of the algorithms used for template extraction, template protection and comparison, as well as all the secrets; possibly execute all the submodules of the system that make use of the secrets. As per the mentioned standard, such conditions would correspond to the worst-case scenario, known as the \textit{full disclosure model}.

\section{BTP Landscape Analysis}
\label{sec:3_Landscape}
In light of the gathered requirements, an analysis of the BTP methods available in the literature was carried out to identify the most suitable solutions. To this end, a useful taxonomy of BTP methods is based on two independent aspects proposed in \cite{btpsurvey}: \textit{(i)} method \textit{type}, which can be \textit{handcrafted} or \textit{NN-based}; \textit{(ii)} method \textit{input}, which can be at \textit{image-level} or at \textit{feature-level}. 

\textbf{Method Type: Handcrafted vs. NN-based}
% \label{subsec:level}

In general terms, the sample(s) or features used as primary input to the BTP scheme are defined as \textit{generative biometric data} \cite{isobtp}. Handcrafted approaches are explicitly defined algorithms applied to generative biometric data to produce protected instances of them. In contrast, NN-based approaches involve training a neural network to learn a suitable protection algorithm, to transform the generative biometric data to a protected template \cite{btpsurvey}. 

NN-based approaches are more recent than handcrafted ones, and they have received attention as they offer the possibility of avoiding explicitly defining the protection method. Nevertheless, the limitations of NN-based methods include the fact that such protection methods are generally specific to the neural network on whose templates they are trained (conflicting with \textit{F.3 -- Feature Extractor-Independence}) \cite{kumar2016deep, kumar2018face}. Moreover, assessing the scalability of these methods without retraining can be difficult (\textit{F.5 -- Easy New Enrolment}) \cite{jami2019biometric}, and the template renewability aspect also appears challenging (\textit{F.6 -- Template Revocability and Renewability}) \cite{pinto2020secure}. 
Concerning the security and privacy analysis, in the adopted full-disclosure threat model, the adversary has access to the trained model (\textit{i.e.}, network architecture and all learned parameters). Consequently, the irreversibility analysis for NN-based BTP methods should consider how this knowledge could be used to extract information about the original embedding or image from different layers of the neural network (\textit{S.1 -- Irreversibility}). Such a thorough irreversibility analysis, which is out of the scope of this work, is still lacking in the literature for these kinds of methods.

\textbf{Method Input: Image-level vs. Feature-level}
% \label{subsec:level}

Image-level methods are applied directly to the biometric sample (\textit{e.g.}, a face image), following which a biometric feature extractor (such as a neural network) would be used to extract features from the protected image. In the case of feature-level methods, a biometric recognition system (most likely a neural network) would first be used to extract a set of features or a ``template" (\textit{i.e.}, an embedding, in the case of a neural-network-based feature extractor) from the biometric sample, then the BTP algorithm would be applied to this template to generate the protected template. 

The overwhelming majority of the works proposed in the literature focuses on applying a BTP method at feature level. This preference for feature-level BTP may be attributed to the availability of several pre-trained NN biometric recognition models, which have been shown to be capable of extracting highly discriminative features from images. In this way, the BTP mechanism could be integrated as an \textit{add-on} module to existing biometric systems, rather than having to additionally design a robust feature extractor for the protected images. To preserve the system modularity, requirement \textit{F.3 -- Feature-Extractor Independence} implies that the BTP method should not work directly on the generative data, allowing us to rule out this category of BTP methods. Additionally, from the perspective of the security and privacy analysis, there is scarcity in the scientific literature concerning the irreversibility and unlinkability properties of these systems (\textit{S.1 -- Irreversibility}, \textit{S.2 -- Unlinkability}), as well as what information about the unprotected template is leaked in different layers of NN-based BTP methods \cite{btpsurvey}. 

\textbf{Feature-level Handcrafted Methods}

We have ruled out NN-based and image-level methods due to their incompatibility with our project requirements. Nevertheless, the majority of BTP methods are both handcrafted and feature-level \cite{btpsurvey}. %Consequently, a number of methods fall into this category. 
In this section, we narrow our focus to three approaches: homomorphic encryption (HE), hashing, and feature transformation approaches (Table \ref{tab:requirements}).

\begin{table}[b!]
\caption{Having narrowed our focus to feature-level handcrafted BTP methods, we consider three approaches: homomorphic encryption, hashing, and PolyProtect \cite{polyprotect}. Below, we roughly rate them against our project requirements and threat model. For the requirements, we consider four possible judgment values: satisfied ($\mathcal{S}$), possible ($\mathcal{P}$), challenging ($\mathcal{C}$), weak ($\mathcal{W}$).}
\label{tab:requirements}
\footnotesize
\centering
\begin{tabular}{|c|c|c|c|c|}
\hline
\multicolumn{2}{|c|}{\textbf{Requirement}} & \makecell[c]{\textbf{HE}} & \makecell[c]{\textbf{Hashing}} & \makecell[c]{\textbf{PolyProtect}\\ \textbf{\cite{polyprotect}}} \\
\hline  
\hline
\textit{F.1} &  \textit{Recognition accuracy} & \makecell[c]{$\mathcal{S}$} & \makecell[c]{$\mathcal{C}$} & \makecell[c]{$\mathcal{S}$} \\
\hline
\textit{F.2} &  \textit{Modality-Independence} & \makecell[c]{$\mathcal{S}$} & \makecell[c]{$\mathcal{P}$} & \makecell[c]{$\mathcal{P}$} \\
\hline
\textit{F.3} & \textit{Feature Extractor-Indep.} & \makecell[c]{$\mathcal{S}$} & \makecell[c]{$\mathcal{P}$} & \makecell[c]{$\mathcal{S}$} \\
\hline
\textit{F.4} & \textit{On-Device Recognition} & \makecell[c]{$\mathcal{C}$} & \makecell[c]{$\mathcal{P}$} & \makecell[c]{$\mathcal{S}$} \\
\hline
\textit{F.5} & \textit{Easy New Enrolment} & \makecell[c]{$\mathcal{S}$} & \makecell[c]{$\mathcal{P}$} & \makecell[c]{$\mathcal{S}$} \\
\hline
\textit{F.6} & \textit{Template Revocability} & \makecell[c]{$\mathcal{S}$} & \makecell[c]{$\mathcal{P}$} & \makecell[c]{$\mathcal{S}$} \\
\hline
\textit{F.7} & \textit{Open-Source} & \makecell[c]{$\mathcal{S}$} & \makecell[c]{$\mathcal{P}$} & \makecell[c]{$\mathcal{S}$}\\
\hline
\hline
\textit{S.1} & \textit{Irreversibility} & \makecell[c]{~$\mathcal{S}$*} & \makecell[c]{$\mathcal{C}$} & \makecell[c]{$\mathcal{S}$} \\
\hline
\textit{S.2} & \textit{Unlinkability} & \makecell[c]{$\mathcal{S}$} & \makecell[c]{$\mathcal{C}$} & \makecell[c]{$\mathcal{S}$} \\
\hline
\hline
\textit{O.1} & \textit{Computational Efficiency} & \makecell[c]{$\mathcal{W}$} & \makecell[c]{$\mathcal{P}$} & \makecell[c]{$\mathcal{S}$} \\
\hline
\textit{O.2} & \textit{Time Effic.} & \makecell[c]{$\mathcal{W}$} & \makecell[c]{$\mathcal{P}$} & \makecell[c]{$\mathcal{S}$} \\
\hline
\textit{O.2} & \textit{Offline Processing} & \makecell[c]{$\mathcal{W}$} & \makecell[c]{$\mathcal{P}$} & \makecell[c]{$\mathcal{S}$} \\
\hline
\hline
\multicolumn{2}{|l|}{\textit{Full-Disclosure Threat Model} \cite{isoperf}} & \makecell[c]{\ding{55}} & \makecell[c]{\textit{?}} & \makecell[c]{\checkmark} \\
\hline
\multicolumn{5}{l}{*Not under the full-disclosure threat model.}\\
\end{tabular}
\end{table}

HE enables us to perform operations on encrypted biometric data without having to first decrypt it, allowing us to maintain the same recognition accuracy, since the comparison score obtained in the encrypted domain is in principle equal to the unprotected system score. Nevertheless, the computational complexity of HE makes it challenging to apply HE as a BTP method (especially in resource-constrained humanitarian contexts).  Consequently, most efforts towards HE-based BTP solutions have focused on reducing this computational complexity while simultaneously trying to minimise the resulting accuracy degradation \cite{boddeti, hers}. In any case, encrypted templates remain secure only insofar as the corresponding decryption key remains secret, conflicting with the full-disclosure threat model adopted in our project.
Hashing operations create a fixed-size, predictable output called a ``hash'', such that it is mathematically impossible to recover the original input data from its hash. In particular, \textit{cryptographic} hash functions are designed to exaggerate small differences in the input. Consequently, the main challenge in their application to biometric templates is due to the intrinsic intra-class variability of biometric samples. For this reason, hash-based BTP methods tend to apply hashing to random, subject-specific codewords, which are bound to the biometric templates by some mathematical function, so that the matching operation takes place indirectly by reconstructing the codewords using probe samples. This is the case for fuzzy committment \cite{fuzzyc} or fuzzy vault-based \cite{fuzzyv} schemes. To reduce sensitivity to intra-class variations, \textit{non-cryptographic} hashes \cite{biohashing} have been proposed: in this case, the same subject's templates are mapped to approximately the same code, allowing distance-based comparisons, without requiring an exact match as for cryptographic hash functions. However, we observed that all hashing methods proposed in the literature are likely to introduce some accuracy degradation \cite{rathgeb2022deep, dong2019cancellable, wang2020biometric} (\textit{F.1 -- Recognition Accuracy}). Moreover, their irreversibility and unlinkability have been demonstrated to be fragile in some cases (\textit{S.1 -- Irreversibility}, \textit{S.2 -- Unlinkability}) \cite{keller2020fuzzy, keller2021inverting}, or non-exhaustively evaluated \cite{kim2021ironmask, dong2021secure}.

Feature transformation approaches are based on transforming templates with the help of subject-specific transformation functions. Although such feature transformations may resemble non-cryptographic hashing methods on the surface (\textit{i.e.}, both are ``transforms'' in a sense), the main difference is that the protected templates generated using feature transformation BTP methods tend to lie in the same (or similar) domain as the original template (\eg, floating-point values), so the same comparison function can often be applied, while hashes tend to be binary, entailing the use of a different comparison function to that adopted in the unprotected template domain (\eg, Hamming distance). 
\textit{PolyProtect}, a method designed for mobile face verification \cite{polyprotect}, falls into this category. The method was developed considering a full disclosure threat model, it shows no significant recognition accuracy degradation, and the code is available under a GPL-3.0 license\footnotemark[3], making the results easily reproducible. 

\footnotetext[3]{\href{https://gitlab.idiap.ch/bob/bob.paper.polyprotect_2021/}{https://gitlab.idiap.ch/bob/bob.paper.polyprotect\_2021/}}

\section{PolyProtect}
\label{sec:4_PolyProtect}
In this section, the adopted feature-transformation BTP method, PolyProtect, is presented. Let $V = [v_1, v_2, ..., v_n]$ be an \textit{n}-dimensional, real-number embedding extracted by a NN. The aim of PolyProtect is to map $V$ to another real-number feature vector, $P = [p_1, p_2, ..., p_k]$ (where $k < n$), which is the protected version of $V$. This is achieved by mapping sets of $m$ (where $m << n$) consecutive elements from $V$ to single elements in $P$ via multivariate polynomials defined by a set of $m$ subject-specific, ordered, unique, non-zero integer coefficients, $C = [c_1, c_2, ..., c_m]$, and exponents, $E = [e_1, e_2, ..., e_m]$. From the perspective of security, the $C$ and $E$ parameters represent secret information. Consequently, they should be securely stored.

The first $m$ consecutive elements of $V$ (\textit{i.e.}, $v_1, v_2, ..., v_m$) are mapped to the first element in $P$ (\textit{i.e.}, $p_1$) via Eq. \ref{eq:1}:
\begin{equation}
    p_{1} = c_{1}v^{e_{1}}_{1} + c_{2}v^{e_{2}}_{2} + ... + c_{m}v^{e_{m}}_{m} 
\label{eq:1}
\end{equation}

The elements of $V$ used to generate $p_2$ depend on the value of overlap $o$ between successive sets of elements. The minimum overlap is 0, in which case the elements of $V$ in each set would be unique. The maximum overlap is $m - 1$, in which case successive element sets would share $m - 1$ elements. Eq. \ref{eq:2} defines the mapping from $V$ to $p_2$ for overlap $o$:

\begin{equation}
p_2 = c_{1}v^{e_{1}}_{m-o+1} + c_{2}v^{e_{2}}_{m-o+2} + ... + c_{m}v^{e_{m}}_{m-o+m} 
\label{eq:2}
\end{equation}

The remaining elements in $P$ (\textit{i.e.}, $p_3, ..., p_k$) are generated in a similar way, until all the elements in $V$ have been used. If the last set in $V$ is incomplete because the dimensionality of $V$ is not divisible by the required number of element sets (defined by $m$ and $o$), $V$ is padded by a sufficient number of zeros to complete the last set.

\begin{figure}[!hb]
 \centering
  \includegraphics[trim={5cm 1.25cm 12.15cm 1.15cm},clip,width=\linewidth]{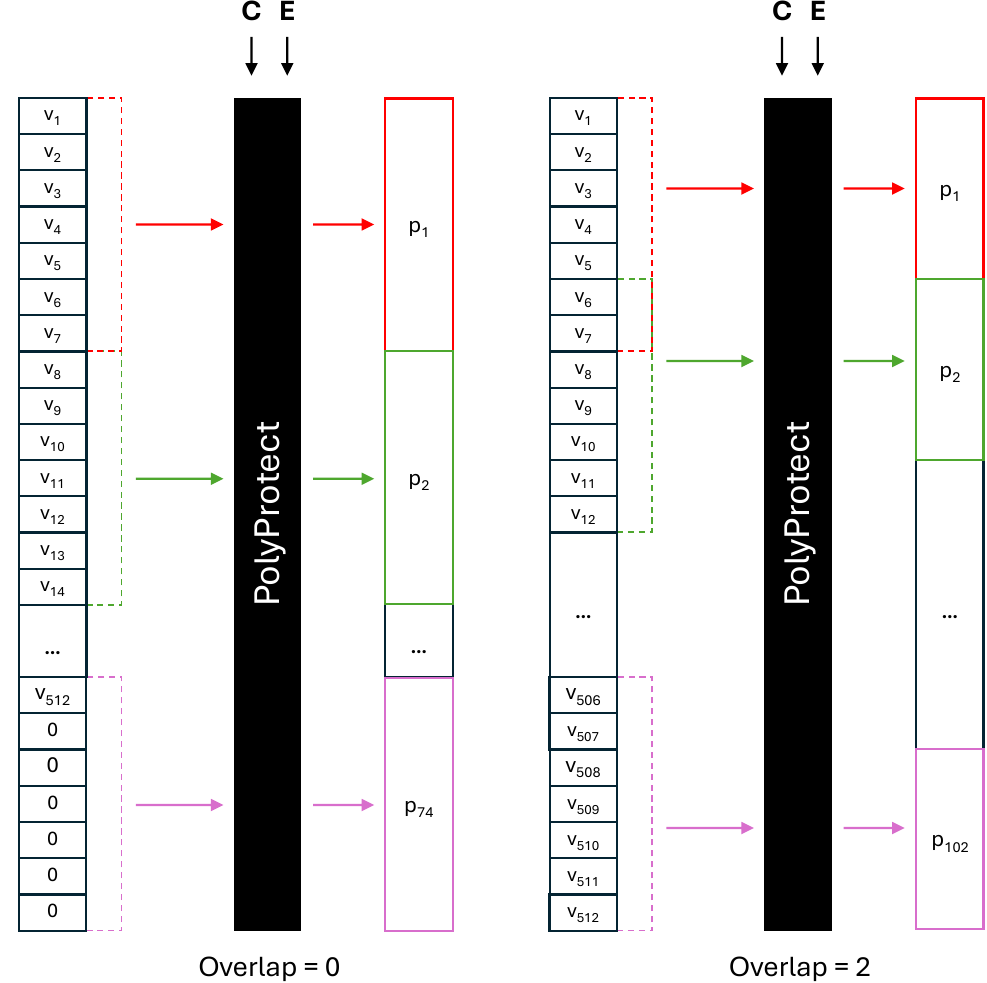}
  \caption{A graphical representation of PolyProtect is provided above. As an example, we consider overlap $o = \{0, 2\}$ for $m = 7$. The dimensionality of the protected template $P$ varies according to both $m$ and $o$.}
  \label{fig:graph_polyprotect}
\end{figure}

\section{Experimental Protocol}
\label{sec:5_Protocol}
In this section, the experimental setup, methodology, and datasets adopted in our experiments are presented.

\subsection{NN-based Feature Extractors}

PolyProtect acts as an additional layer on top of a traditional NN-based biometric recognition system. Consequently, it is first necessary to establish the baseline performance of the corresponding unprotected system, to which the protected system will then be compared. To achieve this, we considered a state-of-the-art efficient face recognition model specifically designed for edge devices called EdgeFace \cite{edgeface}\footnotemark[4]. 
EdgeFace is based on a hybrid network architecture that leverages convolutional NN and Vision Transformer (ViT) capabilities \cite{edgenext}. It produces 512-dimensional output embeddings, which are compared to each other using the cosine distance. 

\footnotetext[4]{We consider the \textit{XS} version, which consists of only 1.77M parameters (6.83MB). Link: \href{https://github.com/otroshi/edgeface}{https://github.com/otroshi/edgeface}}

For fingerprints, preliminary experiments showed the unsuitability of traditional minutiae-based templates for PolyProtect, essentially due to: \textit{(i)} the variable length of minutiae-based template representations, and their inherent two-dimensional nature, as each minutia point is typically described by its location coordinates and type; \textit{(ii)} the lower level of abstraction of minutiae-based templates, in which each minutia point directly corresponds to a feature detected by the sensor (in contrast, NN-based embeddings are produced by a network trained to map identities to a high-dimensional space); \textit{(iii)} the higher complexity of the minutiae-based matching algorithms in comparison with cosine distance (which is typically used for fixed-side NN-based templates). Consequently, despite the greater availability of minutiae-based fingerprint systems, we opted for a pretrained deep learning-based feature extractor\footnotemark[5] \cite{rohwedder2023benchmarking}. The model replicates the DeepPrint architecture \cite{engelsma}, consisting of two main branches: 
one dedicated to the fingerprint texture representation, and the other designated to learning from the minutiae maps. The final 512-dimensional embedding consists of the concatenation of the outputs of the two branches, \textit{i.e.}, the first half of the embedding encodes fingerprint texture information and the second half encodes minutiae information. 

\footnotetext[5]{\href{https://github.com/tim-rohwedder/fixed-length-fingerprint-extractors}{https://github.com/tim-rohwedder/fixed-length-fingerprint-extractors}} 

\subsection{Configuration of PolyProtect}

Implementing PolyProtect requires setting four values: $m$ (the number of unprotected template elements used to compute each element of the protected template $P$), $o$ (the amount of overlap between consecutive sets of $m$ elements), and the ranges of $C$ (coefficients of the polynomial) and $E$ (exponents of the polynomial). 

The minimum value of $m$ is set by the Abel-Ruffini theorem \cite{polyprotect}, which states that there is no closed form algebraic expression for solving polynomials of degree 5 or higher with arbitrary coefficients \cite{gray2018unsolvability}. At the same time, $m$ corresponds to the length of the secret sequences of coefficients $C$ and exponents $E$. In particular, the $E$ values are selected as random permutations of integers from 1 to $m$. Therefore, to avoid having two enrolled subjects with the same set of exponents, for $m = 5$ we would have 120 subjects, for $m = 6$ we would have 720, for $m = 7$ we would have 5040, etc. 
The upper limit of $m$ is bound by the fact that the embeddings produced by the NN-based feature extractors consist of floating point values smaller than 1, which would cause large powers to effectively obliterate certain elements. In \cite{polyprotect}, the authors of PolyProtect set $m$ to 5, since their baseline evaluation is carried out on a dataset consisting of fewer than 100 subjects.
Accounting for bigger datasets, we set $m = 7$. 
Concerning the range of the coefficients $C$, the value adopted in the original paper was arbitrarily set to $[-50, 50]$. We extend this range to $[-100, 100]$ to compensate for the increase of $m$ from 5 to 7. 

\subsection{Analysis of BTP Criteria}

To reproduce the original experiments, the subjects are split into \textit{dev} and \textit{eval} sets (in equal proportions), each subject having a single reference and query sample image. Following the original paper, we consider two scenarios for the verification accuracy evaluation: \textit{(i)} the \textit{Normal} (N) scenario, in which the system should operate most of the time; and \textit{(ii)} the \textit{Stolen Coefficients and Exponents} (SCE) scenario, in which a subject attempts to authenticate as a different one by stealing the target's $C$ and $E$ parameters, and applying them to their own embedding to generate their PolyProtected template. 
Concerning the identification scenario, given the absence of the identity claim, it would be impossible to know which subject-specific parameters to use for the transformation. Therefore, the query template must be transformed by PolyProtect considering all sets of transformation parameters ($C$ and $E$) already registered in the database. Then, a ranking is generated based on the distances, and the highest-scoring matches are returned. 

In our irreversibility analysis, the full disclosure threat model described in Sec. \ref{sec:2_requirements} is reflected as follows: knowledge of the algorithm including the number of embedding elements ($m$) used to generate each PolyProtected element, the overlap value ($o$), as well as the subject-specific $C$ and $E$ parameters. Moreover, we assume that the adversary has access to one or more PolyProtected templates, $P$, corresponding to a particular embedding, $V$, as well as knowledge of the distribution of unprotected templates, which is representative of the embeddings used to create the PolyProtected templates to which the adversary has access. The adversary's goal, therefore, is to use all this information to attempt to recover a subject's original embedding, $V$, from one or more of their PolyProtected templates, $P$. Specifically, the embeddings to be recovered are the evaluation set reference embeddings, which the adversary does not have access to. In contrast, we assume that the adversary has access to the development set, which is used to estimate the distribution of each one of the 512 values in the evaluation set reference embeddings as well as the match threshold. Following the attack defined in the original paper, a numerical solver\footnotemark[6] starts from guesses obtained from the development set to estimate a solution for each of the 512 elements in each $V$ in the evaluation set, from the corresponding $P$. \footnotetext[6]{The numerical solver used is Python's \textit{scipy.optimize.root} function with the \textit{lm} method.\\ Link: \href{https://docs.scipy.org/doc/scipy/reference/optimize.root-lm.html}{https://docs.scipy.org/doc/scipy/reference/optimize.root-lm.html}}
We also consider an Attack via Record Multiplicity (ARM): the adversary has access to multiple PolyProtected templates from the same $V$, which they attempt to combine to recover an approximation of $V$. This type of attack could occur in the scenario where the same embedding is used to generate different PolyProtected templates (using different $C$ and $E$ parameters), then each PolyProtected template is either enrolled in a different application or used to replace a compromised PolyProtected template in the same application. 
This was simulated using the same numerical solver approach, but considering $k \times p$ equations (where $k$ is the dimensionality of the $P$ templates, and $p$ is the number of $Ps$ that the adversary is assumed to have access to), instead of only $k$ equations. 
We invite the reader to consult \cite{polyprotect} for more details about the definition of the system equations, which we omit for brevity. 

The unlinkability of PolyProtect is evaluated using the framework proposed in \cite{unlink}. This framework is based on mated and non-mated score distributions, which represent the comparison scores between different protected templates from the same subject and between different protected templates from different subjects, respectively. The unlinkability is measured in terms of $D_{\leftrightarrow}^{sys}$, a global measure of the overall linkability of the underlying recognition system.
Following the recommendation in \cite{unlink}, we compute 10 different PolyProtected templates per person. 
Then, each PolyProtected template is compared to every other PolyProtected template from the same subject to generate a set of mated comparison scores, and to all PolyProtected templates from every other subject to generate a set of non-mated comparison scores. We also calculate the unlinkability of the corresponding unprotected embeddings in the same way.

\subsection{Datasets}

For face biometrics we employ a dataset of face images obtained within the framework of a project currently being carried out in Ethiopia. It consists of 942 subjects with 2 captures per subject: 57\% of the subjects are females, and 43\% are males, while the mean age is approximately 24.5 years ($\sigma = 16.5$). All subjects have East African origins. 
For fingerprint biometrics, we adopt an internal dataset collected in a field project in Ghana, consisting of 119 subjects with two samples each. 
Images were mostly acquired outdoors, from collaborative subjects, by trained data collectors. For face, low-end smartphones were used, with 8-12MP resolution. For fingerprints, dedicated scanners were used, with a resolution of 500DPI. Dirt and dust often accumulated on the scanner surface, making the capture challenging. 
The datasets cannot be made public due to participant privacy agreements, but the code will be made public so that interested researchers can perform evaluations on their own datasets of interest.

\section{Experimental Results}
\label{sec:6_Results}
This section presents our experimental results in terms of recognition accuracy, irreversibility, and unlinkability.

\subsection{Recognition Accuracy}
\label{subsec:ra_res}

\begin{table}[b]
\caption{Verification results. In bold, the results achieved by PolyProtect which improve the baseline performance.}
\label{tab:results}
\centering
\footnotesize
\begin{tabular}{|p{0.5cm}|p{1.2cm}|p{1.15cm}|p{1.15cm}|p{1.15cm}|}
\hline
\makecell[c]{$o$} & \makecell[c]{TMR (\%)\\@ FMR\\= 0.01\% $\uparrow$} & \makecell[c]{TMR (\%)\\@ FMR\\= 0.1\% $\uparrow$} & \makecell[c]{EER\\ (\%) $\downarrow$} & \makecell[c]{EER\\ (\%) $\downarrow$} \\
\hline
\multicolumn{1}{c|}{} & \multicolumn{3}{c|}{Face} & \multicolumn{1}{c|}{Fingerprint} \\  
\hline
\makecell[c]{Bas.} & \makecell[c]{94.27} & \makecell[c]{95.97} & \makecell[c]{1.35}  & \makecell[c]{40.21} \\
\hline
\multicolumn{5}{|c|}{N Scenario} \\
\cline{1-5}
\makecell[c]{0} & \makecell[c]{93.40} & \makecell[c]{\textbf{97.26}} & \makecell[c]{\textbf{0.83}}  & \makecell[c]{\textbf{7.13}}\\
\cline{1-5}
\makecell[c]{1} & \makecell[c]{\textbf{94.67}} & \makecell[c]{\textbf{97.24}} & \makecell[c]{\textbf{0.72}}  & \makecell[c]{\textbf{6.82}} \\
\cline{1-5}
\makecell[c]{2} & \makecell[c]{\textbf{94.78}} & \makecell[c]{\textbf{98.07}} & \makecell[c]{\textbf{0.57}}  & \makecell[c]{\textbf{6.69}}\\
\cline{1-5}
\makecell[c]{3} & \makecell[c]{\textbf{94.97}} & \makecell[c]{\textbf{97.71}} & \makecell[c]{\textbf{0.65}}  & \makecell[c]{\textbf{6.51}}\\
\cline{1-5}
\makecell[c]{4} & \makecell[c]{\textbf{95.65}} & \makecell[c]{\textbf{98.51}} & \makecell[c]{\textbf{0.62}}  & \makecell[c]{\textbf{6.19}}\\
\cline{1-5}
\makecell[c]{5} & \makecell[c]{\textbf{95.97}} & \makecell[c]{\textbf{98.54}} & \makecell[c]{\textbf{0.55}}  & \makecell[c]{\textbf{6.30}} \\
\cline{1-5}
\makecell[c]{6} & \makecell[c]{\textbf{96.39}} & \makecell[c]{\textbf{98.98}} & \makecell[c]{\textbf{0.45}}  & \makecell[c]{\textbf{6.19}}\\
\cline{1-5}
\multicolumn{5}{|c|}{SCE Scenario}\\
\cline{1-5}
\makecell[c]{0} & \makecell[c]{87.54} & \makecell[c]{92.80} & \makecell[c]{2.25} & \makecell[c]{40.83}\\
\cline{1-5}
\makecell[c]{1} & \makecell[c]{89.38} & \makecell[c]{93.95} & \makecell[c]{2.21}  & \makecell[c]{40.68}\\
\cline{1-5}
\makecell[c]{2} & \makecell[c]{90.74} & \makecell[c]{93.99} & \makecell[c]{1.74}  & \makecell[c]{41.73}\\
\cline{1-5}
\makecell[c]{3} & \makecell[c]{90.28} & \makecell[c]{94.59} & \makecell[c]{1.91}  & \makecell[c]{40.5}\\
\cline{1-5}
\makecell[c]{4} & \makecell[c]{91.95} & \makecell[c]{95.05} & \makecell[c]{1.51}   & \makecell[c]{40.64}\\
\cline{1-5}
\makecell[c]{5} & \makecell[c]{93.36} & \makecell[c]{95.76} & \makecell[c]{1.49}  & \makecell[c]{\textbf{39.95}}\\
\cline{1-5}
\makecell[c]{6} & \makecell[c]{94.20} & \makecell[c]{\textbf{96.05}} & \makecell[c]{\textbf{1.26}}  & \makecell[c]{40.34}\\
\hline
\end{tabular}
\end{table}

The verification performance is measured in terms of True Match Rate (TMR) at a given False Match Rate (FMR). We set thresholds corresponding to FMR=0.01\% and 0.1\%, as well as to the Equal Error Rate (EER).  
Table \ref{tab:results} shows that for almost any overlap value, PolyProtect improves the verification performance with respect to the baseline performance (except for TMR at FMR=0.01\%, with an overlap of 0). 
This might be due to the fact that by combining subject-specific information ($C$ and $E$ parameters), in the protected space embeddings belonging to the same subject are pushed together, while embeddings belonging to different subjects are moved away from each other. 
On top of this, it is clear that by increasing the overlap value (down the rows), the performance improves.  
PolyProtect always reduces the dimensionality of the input template. In particular, the higher the overlap value, the higher the number of dimensions of the protected space (from 512 values of the input unprotected template, for an overlap of 6 the output template will have a dimensionality of 506, while for an overlap of 0 the dimensionality of the output template will be 74). So, it makes sense that the use of larger overlap values, which generate PolyProtected templates of higher dimensionality, will result in higher recognition accuracy.

The bottom part of Table \ref{tab:results} contains the results in the SCE scenario. A reduction of the biometric performance is expected \cite{polyprotect}, as embeddings transformed with the same secret parameters are being compared. Indeed, in contrast to the N scenario, the baseline performance is, in almost all cases, the best one (except for TMR at FMR=0.1\% and EER, with an overlap of 6). 
Moreover, similarly to the N scenario, the performance improves with an increase in the overlap value, thus limiting the accuracy degradation.

For fingerprint biometrics, we observe a baseline performance of 40.21\% EER. As mentioned in Sec. \ref{sec:5_Protocol}, we employ pretrained models without any fine tuning. Due to the worse model performance (compared to the face recognition system), in this case we omit the results at stricter operating points such as FMR = 0.1\%. With a lower baseline recognition accuracy, in the N scenario the impact of PolyProtect seems to be much greater, reducing the EER to 7.13\% for an overlap of 0. In field operations, where, due to harsh operational conditions, baseline performance levels typically tend to be lower than on datasets assembled in laboratory settings, this kind of contribution could be of great added value. Moreover, interestingly, we notice that increasing the overlap value (down the rows) does not yield a clear improvement trend as in the face experiment. 

Additionally, we observe that, as in the case of face biometrics, the EER values obtained in the SCE scenario are almost always higher than in the baseline system. The performance discrepancy between the N and SCE scenarios seems to confirm the role played by the subject-specific secret information ($C$ and $E$ parameters) towards producing a more discriminative mapping in the protected domain.

\begin{table}[b!]
\caption{Identification results in terms of True Positive Identification Rate-$n$ (TPIR-$n$), which represents the percentage of identification attempts where the query subject is included in the ranked list of the $n$ most similar candidates returned after searching a biometric reference database.}
\centering
\footnotesize
\begin{tabular}{|c|c|c|c|c|c|c|}
\hline
\multirow{2}{*}{$o$} & \multicolumn{3}{c|}{Face (TPIR-\textit{n} (\%), \textit{n})} & \multicolumn{3}{c|}{Fingerprint (TPIR-\textit{n} (\%), \textit{n})} \\
\cline{2-7}
& n = 1 & n = 3 & n = 10 & n = 1 & n = 3 & n = 10 \\
\hline
Bas. & 98.09 & 98.51 & 99.36 & 11.67 & 23.33 & 41.67 \\ \hline
0 & 93.74 & 96.16 & 97.92 & 8.00 & 19.83 & 37.83 \\ \hline
1 & 95.10 & 96.62 & 98.03 & 8.50 & 18.67 & 38.33 \\ \hline
2 & 95.78 & 97.37 & 98.66 & 7.50 & 18.33 & 39.17 \\ \hline
3 & 95.78 & 97.37 & 98.41 & 8.33 & 18.83 & 41.00 \\ \hline
4 & 96.47 & 97.86 & 98.88 & 9.50 & 18.17 & 39.33 \\ \hline
5 & 97.39 & 97.96 & 98.85 & 8.67 & 21.33 & 41.17 \\ \hline
6 & 97.41 & 98.47 & 99.19 & 10.17 & 19.67 & 41.00 \\ \hline
\end{tabular}
\label{tab:id_results}
\end{table}

In the identification scenario, given the absence of an identity claim, the query template must be transformed by PolyProtect considering all sets of transformation parameters ($C$ and $E$) registered in the database. All the protected query templates obtained are then compared to the corresponding protected reference template, \textit{i.e.}, the one transformed with the same set of $C$ and $E$ parameters during enrolment. Given this aspect, identification is inevitably more difficult than verification. In fact, the comparisons always take place between pairs of protected templates that undergo exactly the same transformation, without exploiting the discriminative power of incorporating subject-specific information. From this perspective, the comparisons are comparable to those carried out in the SCE verification scenario.

Table \ref{tab:id_results} shows that for face the baseline performance is better in all cases. However, by increasing the overlap value, and therefore the dimensionality of the output (protected) space, we observe an overall improvement in the identification rates. 
For TPIR-3 and TPIR-10, Table \ref{tab:id_results} shows that for intermediate overlap values, such as 2 or 3, the decrease in identification accuracy when PolyProtect is employed is approximately 1\% in absolute terms, which would be considered acceptable. For the more stringent TPIR-1, the corresponding decrease would be greater (approximately 2.5\% in Table \ref{tab:id_results}). 
For fingerprints, the adopted dataset proves to be very challenging in the identification task as well, but the negative impact of PolyProtect seems limited in this case. 

\begin{table}[t!]
 \caption{Irreversibility results in terms of Inversion Success Rate (ISR).}
\label{tab:inv_single}
 \centering
\footnotesize	
\begin{tabular}{|c|c|c|c|c|}
\hline
\multirow{2}{*}{$o$} & \multicolumn{2}{c|}{Face (ISR (\%))} & \multicolumn{2}{c|}{Fingerprint (ISR (\%))} \\
\cline{2-5}
 & FMR = 0.01\% & FMR = 0.1\% & FMR = 1\% & FMR = 10\% \\ \hline
0 & 0 & 0 & 0 & 1.00 \\ \hline
1 & 0 & 0 & 0 & 1.50 \\ \hline
2 & 0 & 0 & 0 & 4.50 \\ \hline
3 & 0 & 0 & 0 & 12.00 \\ \hline
4 & 0 & 0.02 & 0 & 37.33 \\ \hline
5 & 0.83 & 34.59 & 9.83 & 79.83 \\ \hline
6 & 98.20 & 98.20 & 92.33 & 92.33 \\ \hline
\end{tabular}
\end{table}

\subsection{Irreversibility}
\label{subsec:irr_res}

Table \ref{tab:inv_single} shows the results of the attempts to reconstruct an unprotected template from a single protected template.  
The Inversion Success Rate (ISR) is computed as the solution rate $\times$ match rate \cite{polyprotect}. 
It is evident that the inversion success rate is, in general, lower when the baseline systems operate at a stricter match threshold (at a lower FMR), since a stricter threshold would require a better approximation of the original input template. 
Another interesting observation is that, as the overlap value increases, the ISR increases, similarly to the recognition accuracy, revealing a trade-off between these two metrics. For the face dataset, it takes at least an overlap of 5 at the stricter threshold to observe 0.83\% of successful reconstructions. Then, the ISR reaches almost 100\%  for an overlap of 6. With the more lenient threshold, we observe a 34.59\% ISR with an overlap of 5. This trend (\textit{i.e.}, increasing ISR as the amount of overlap increases) is due to the number of equations in the underdetermined system of equations assembled for attempting the reconstruction starting from a single template, \textit{i.e.}, the greater the overlap, the greater the number of equations, and the more constrained the system becomes, so it becomes easier to solve for $V$ \cite{polyprotect}. 

\begin{figure}[t!]
 \centering
   \centering
   \includegraphics[width=0.43\textwidth]{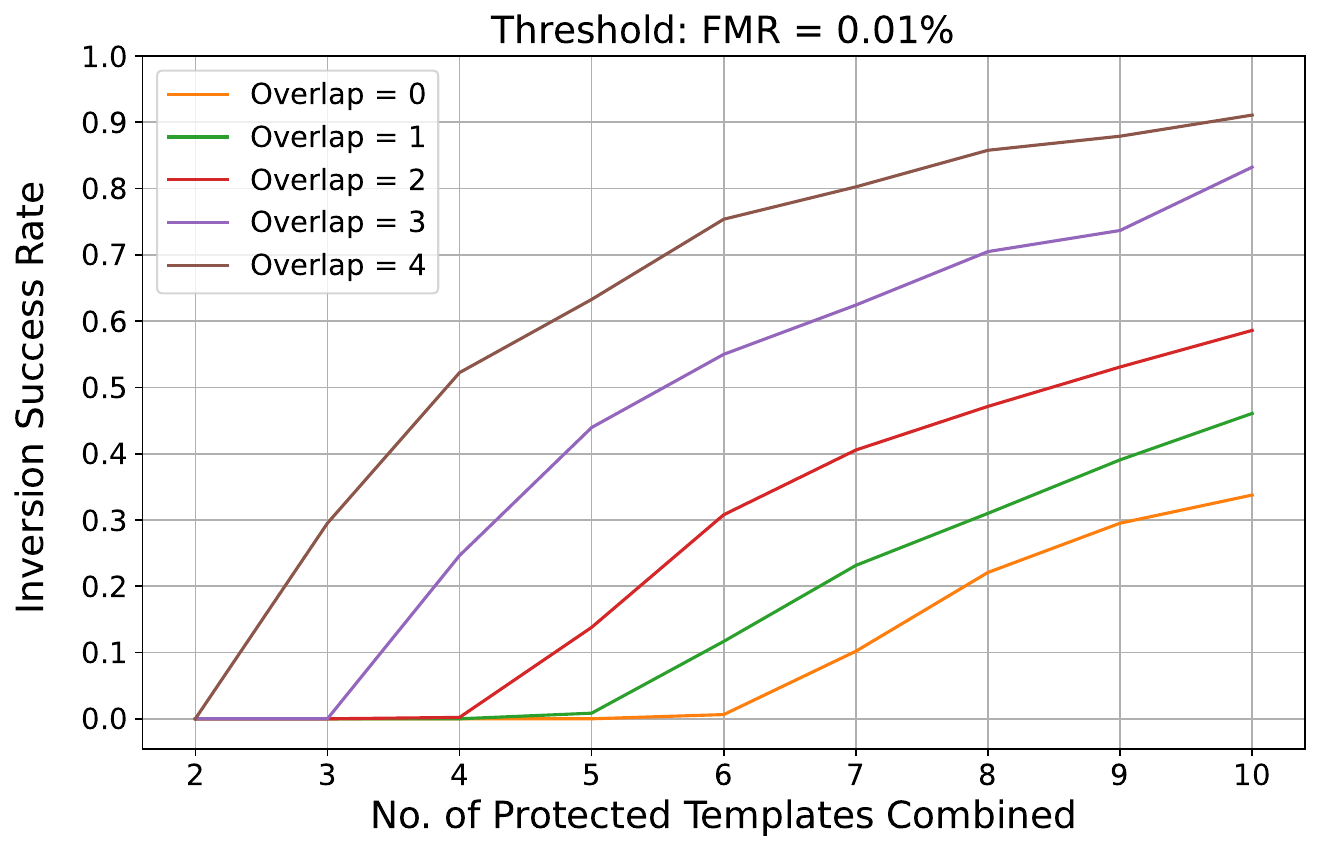}
   \caption{Attack via Record Multiplicity (ARM) for the face protected system.}
 \label{fig:arm}
\vspace{-0.5cm}
\end{figure}

Fig. \ref{fig:arm} shows the results of the ARM experiment on the face dataset. 
Our ARM analysis encompasses a maximum overlap value of 4 since we observed non-zero ISR values for an overlap of 5 in the inversion of a single template.
The used protected templates were all generated from the same embedding but using different $C$ and $E$ parameters. The number of equations in the system to be solved by the numerical solver consists of the number of protected templates that are combined $\times$ the dimensionality of each protected template. In turn, the number of unknowns would be equal to the dimensionality of each unprotected template. Overall, we can observe that for a given overlap value, as the number of combined protected templates increases, the chances of a successful reconstruction are higher. 

\subsection{Unlinkability}
\label{subsec:unl_res}

Table \ref{tab:unlink} summarises the global $D_{\leftrightarrow}^{sys}$ measures for all overlaps. Concerning the metrics adopted, $D_{\leftrightarrow}^{sys}$ measures the overall system linkability, where a value of 0 would indicate that the system is fully unlinkable, whereas a value of 1 would indicate that the system is fully linkable \cite{unlink}. We observe that $D_{\leftrightarrow}^{sys}$ for our baseline systems, which use unprotected face embeddings, is closer to 1 (or at least significantly further from 0 compared to the protected systems). 

Compared to the naive (random) parameter selection, the strict one involves an additional comparison check \cite{polyprotect}: sets of $C$ and $E$ parameters are selected only if they are capable of producing a protected template $P$ such that the comparison scores with all the other protected templates originating from the same unprotected template $V$, obtained with the other assigned sets of $C$ and $E$ parameters, are within a required score range. Otherwise, a new set of parameters is randomly generated until the aforementioned condition is satisfied. The idea behind this strict process of selecting the $C$ and $E$ parameters is to ensure that different protected templates generated from the same face embedding would be unlinkable, \textit{i.e.}, comparison of these templates would generate scores in the ``unlinkable" score range. 
In contrast to the unprotected systems, the $D_{\leftrightarrow}^{sys}$ values for our protected systems are reduced by a factor of 10, and are close to 0, with the naive parameter selection, suggesting that different protected templates generated from the same subject’s face or fingerprint embedding are almost fully unlinkable. The strict PolyProtect parameter selection shows that the $D_{\leftrightarrow}^{sys}$ values are further reduced, especially in the case of the fingerprint dataset. 

\begin{table}[h!]
\caption{Unlinkability results, considering the naive (N) and strict (S) PolyProtect parameter ($C$ and $E$) selection.}
\footnotesize
\label{tab:unlink}
\centering
\begin{tabular}{|c|c|c|c|c|}
% {|p{1cm}|p{0.8cm}|p{0.8cm}|p{0.8cm}|p{0.8cm}|}
\cline{2-5}
\multicolumn{1}{c|}{} & \multicolumn{4}{c|}{$D^{sys}_{\leftrightarrow}$} \\
\cline{2-5}
\multicolumn{1}{c|}{} & \multicolumn{2}{c|}{\makecell[c]{Face}} & \multicolumn{2}{c|}{\makecell[c]{Fingerprint}}\\
\cline{2-5}
\hline
$o$ & \makecell[c]{N} & \makecell[c]{S} & \makecell[c]{N} & \makecell[c]{S}\\
\hline
\makecell[c]{Bas.} & \multicolumn{2}{c|}{\makecell[c]{0.757}} & \multicolumn{2}{c|}{\makecell[c]{0.277}} \\
\hline
\makecell[c]{0} & \makecell[c]{0.077} & \makecell[c]{0.062} & \makecell[c]{0.092} & \makecell[c]{0.022}\\
\cline{1-5}
\makecell[c]{1} & \makecell[c]{0.078} & \makecell[c]{0.063} & \makecell[c]{0.089} & \makecell[c]{0.027}\\
\cline{1-5}
\makecell[c]{2} & \makecell[c]{0.078} & \makecell[c]{0.065} & \makecell[c]{0.095} & \makecell[c]{0.022} \\
\cline{1-5}
\makecell[c]{3} & \makecell[c]{0.078} & \makecell[c]{0.065} & \makecell[c]{0.089} & \makecell[c]{0.033}\\
\cline{1-5}
\makecell[c]{4} & \makecell[c]{0.078} & \makecell[c]{0.069} & \makecell[c]{0.097} & \makecell[c]{0.028} \\
\cline{1-5}
\makecell[c]{5} & \makecell[c]{0.078} & \makecell[c]{0.072} & \makecell[c]{0.089} & \makecell[c]{0.034} \\
\cline{1-5}
\makecell[c]{6} & \makecell[c]{0.078} & \makecell[c]{0.077} & \makecell[c]{0.099} & \makecell[c]{0.039}\\
\hline
\end{tabular}
\end{table}

\vspace{-0.5cm}
\section{Conclusions}
\label{sec:7_Conclusions}
This article presented a BTP-enhanced mobile biometric system, suitable for humanitarian and emergency scenarios, which are characterised by particularly strict functional, operational, and security and privacy requirements (formulated in Sec. \ref{sec:2_requirements}), as an unsafe use of biometrics can be particularly harmful for the data subjects.
We provided a broad analysis of the existing BTP literature to identify a suitable method for our requirements (Sec. \ref{sec:3_Landscape}). 
We then selected PolyProtect \cite{polyprotect}, a feature-transformation approach, over approaches such as HE and hashing, mainly due to: its extremely lightweight computational burden; its property of using the same mathematical operation for matching in the unprotected and protected space (\textit{i.e.}, cosine distance); and its suitability for the full disclosure threat model \cite{isoperf}. 

We experimentally validated PolyProtect, observing results consistent with prior findings in recognition accuracy, irreversibility, and unlinkability. The novel insights provided in this work can be summarised as follows: 
\textit{(i)} we configured our system to deal with a higher number of enrolled subjects (hence $m = 7$) and tested it on a challenging face dataset collected in a field project in Ethiopia, with very promising results, in combination with a more recent and efficient feature extractor, EdgeFace; \textit{(ii)} we extended the evaluation of the recognition performance to the task of identification, showing that the performance degradation with respect to the unprotected system would probably be acceptable in practice; \textit{(iii)} we tested PolyProtect on fingerprint embeddings from a dataset collected in Ghana, considering a fixed-length, freely available implementation of DeepPrint \cite{engelsma, rohwedder2023benchmarking}, showing for the first time the true cross-modality potential of this BTP method.

{\small
\bibliographystyle{ieee}
\bibliography{egbib}
}

\end{document}